\documentclass[runningheads]{llncs}
\usepackage[T1]{fontenc}
\usepackage{graphicx,subfigure}
\usepackage{amssymb}
\usepackage{biblatex}
\usepackage{hyperref}
\usepackage{xcolor}
\usepackage{url}
\begin{document}

\title{Adversarial Data Poisoning for Fake News Detection: How to Make a Model Misclassify a Target News without Modifying It\thanks{This work was partially supported by projects FAIR (PE0000013) and SERICS (PE00000014) under the MUR National Recovery and Resilience Plan funded by the European Union - NextGenerationEU. Supported also by the ERC Advanced Grant 788893 AMDROMA,  EC H2020RIA project “SoBigData++” (871042), PNRR MUR project  IR0000013-SoBigData.it.}}
\titlerunning{Targeted Data Poisoning in Misinformation}

% Federico Siciliano
% Luca Maiano
% Lorenzo Papa
% Irene Amerini
% Fabrizio Silvestri

\author{
Federico Siciliano\inst{1}\orcidID{0000-0003-1339-6983} \and
Luca Maiano\inst{1}\orcidID{0000-0001-7969-7821} \and
Lorenzo Papa\inst{1}\orcidID{0000-0002-9393-5248} \and
Federica Baccini\inst{1}\orcidID{0000-0001-6107-9611} \and
Irene Amerini\inst{1}\orcidID{0000-0002-6461-1391} \and
Fabrizio Silvestri\inst{1}\orcidID{0000-0001-7669-9055}
}
\authorrunning{F. Siciliano, L. Maiano, L. Papa, F. Baccini, I. Amerini, F. Silvestri}
\institute{
    $^1$Sapienza University of Rome, Rome, Italy
    \email{\{siciliano,maiano,papa,baccini,amerini,fsilvestri\}@diag.uniroma1.it}
    }
\maketitle

\begin{abstract}
\looseness -1 Fake news detection models are critical to countering disinformation but can be manipulated through adversarial attacks.
%in which malicious actors tamper with training data. 
In this position paper, we analyze how an attacker can compromise the performance of an online learning detector on specific news content without being able to manipulate the original target news.
In some contexts, such as social networks, where the attacker cannot exert complete control over all the information, this scenario can indeed be quite plausible.
Therefore, we show how an attacker could potentially introduce poisoning data into the training data to manipulate the behavior of an online learning method.
Our initial findings reveal varying susceptibility of logistic regression models based on complexity and attack type.

%Fake news detection models play a crucial role in combating the spread of misinformation. However, these models can be susceptible to adversarial attacks, where malicious actors manipulate the training data to deceive the model's classification decision.
%This paper investigates the concept of adversarial data poisoning in fake news detection models, specifically focusing on methods that can make a model misclassify a true news article as false without directly modifying the target article.
%We illustrate the literature on the subject by pointing out that no one has dealt with the problem of data poisoning by Misinformation Detection from our proposed perspective.
%We then list possible datasets, models, and poisoning methods that can be used to study the scenario we present.
%Preliminary results show that the susceptibility of logistic regression models varies based on their complexity and the poisoning attack type.
%Finally, we discuss future work, including implementing the proposed methods, proposing countermeasures against poisoning attacks, and exploring poison recognition techniques.
%This research sheds light on the risks associated with adversarial data poisoning and highlights the need for robust defenses to protect fake news detection models in the face of evolving adversarial threats.

\end{abstract}

\section{Introduction}

\looseness -1 AI plays a crucial role in recognizing fake news online. Indeed, automated verification is indispensable in the fight against the dissemination of misleading content, especially in the context of large social platforms.
%\footnote{\url{https://ai.meta.com/blog/the-shift-to-generalized-ai-to-better-identify-violating-content/}}
In this scenario, detectors should be designed to continuously learn to classify recent news without affecting the performance obtained from previously acquired knowledge. As a consequence, online learning plays a crucial role in the design of such models~\cite{9900151}. In this dynamic framework, adversarial attacks can compromise the performance of detectors on some news items.

\looseness -1 This position paper aims to explore the concept of adversarial data poisoning~\cite{9900151} in the context of online learning fake news detectors. Specifically, we investigate suitable methods to manipulate a model to ultimately misclassify a true news article as false \emph{without} directly modifying the target article. This type of manipulation reflects the realistic scenario in which the attacker cannot control all the spreading news but can deliberately attack the detector to make it misclassify a specific news item by introducing new poisoned examples. %%%%%

\looseness -1 Prior research has primarily focused on modifying the target news articles~\cite{PRICE2019175,campanile2021vulnerabilities} to manipulate model behavior. This approach poses practical limitations and requires direct access to the articles.
Moreover, while many studies concentrate on recognizing fake news in an offline scenario~\cite{10.1145/3395046}, Horne et al.~\cite{10.1145/3363818} show that traditional content-based methods' performances slowly degrade over time, requiring periodic retraining, which can be mitigated through online learning procedures. Despite the robustness of this learning method, a few studies have considered applying online learning to content-based fake news detection methods~\cite{10.1145/3363818,10.1145/3472619}.

\looseness -1 Unlike previous works, by means of careful selection and incorporation of adversarial examples into the training data, we seek to understand the vulnerabilities of the online learning model~\cite{hoi2021online} and the potential risks associated with data poisoning attacks. By shifting the focus to modify the training data without altering the target article, we aim to explore a more covert and scalable form of attack, highlighting the need for robust defenses against data poisoning techniques. 

%The paper is organized as follows. Section~\ref{sec::method} presents the online learning framework. Section~\ref{sec::attacks} describes two types of data poisoning attacks for the proposed framework, while Section~\ref{sec::results} discusses preliminary results on the analysis of two logistic regressors. Finally, in Section~\ref{sec::concl}, we draw our conclusions and share some hints on possible future research directions.
%\textcolor{red}{We hope this work will attract the community's attention in this direction and stimulate discussion through this initial preliminary analysis of the problem.}
%\subsection{Problem}
%The problem at hand is the susceptibility of online learning models, particularly those used for fake news detection, to adversarial data poisoning attacks. 

%\subsection{Solution / Method}
%Our proposed method revolves around the strategic incorporation of carefully selected adversarial examples into the training data of an online fake news detection model. By exploiting the model's vulnerabilities and subtly manipulating the learning process, we intend to investigate how the model's classification decision can be influenced to misclassify a true news article as false. %This approach avoids the need for direct modification of the target news article, making it potentially more challenging to detect and mitigate.

%\subsection{Outcome}
%The anticipated outcome of this research is to shed light on the potential threats posed by adversarial data poisoning in the online learning context, specifically within the domain of fake news detection. 
\looseness -1 %By demonstrating the feasibility of manipulating model behavior without modifying the target news article, 

\section{Online learning framework}\label{sec::method}

%\subsection{Online Learning and Data Poisoning}

\begin{figure}[t]
\includegraphics[width=\textwidth]{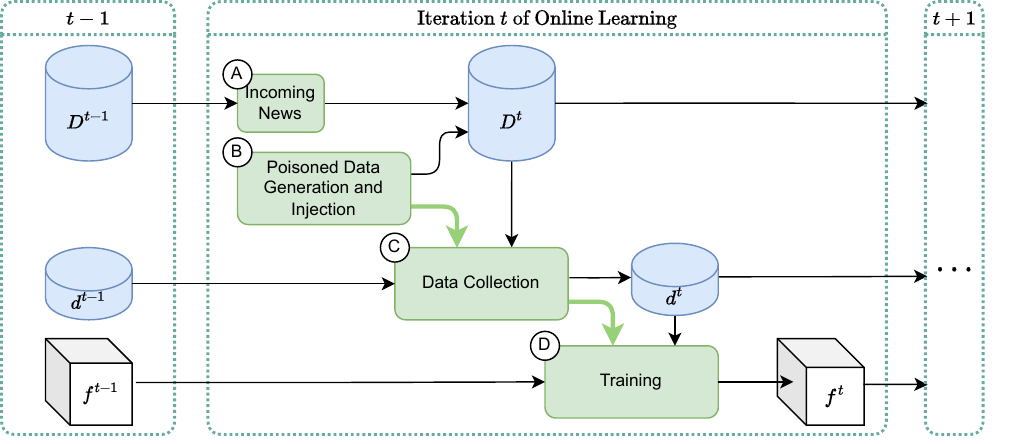}
\caption{Generic iteration at time $t$ of the iterative process of online learning with data poisoning. \textbf{A}: new incoming news adds up to the already existing ones ($D^{t-1}\rightarrow D^t$). \textbf{B}: poisoned data are generated and injected into the existing data. \textbf{C}: a subset of the data is collected and added to the data $d^{t-1}$ collected at time $(t-1)$; the aggregated data are denoted as $d^t$. \textbf{D}: the model $f^{t-1}$ is updated to $f^t$ with the addition of the newly collected data.}\label{fig:process}
\end{figure}
In this section, we describe a realistic online learning framework in which poisoned data are injected into the training data to misguide a model's prediction. Figure~\ref{fig:process} depicts the iterative process of an online learning fake news detector. As shown in the figure, at each time $t$, previously unseen news articles are combined with already existing ones; then, some poisoned data are generated and added to the collected data. Finally, the model is trained and updated using all the aggregated data. 
Since online learning involves the fake news detection model actively collecting news articles at different time instants, the model remains up-to-date and adapts to the ever-changing landscape of online information. Moreover, the framework not only encompasses the new data, but also integrates historical data previously encountered. 

Within this dynamic scenario, it is possible to analyze the effects of the deliberate introduction of poisoned samples aimed at challenging the model's ability to distinguish between genuine and fake news. To support this observation, in the next section, we report experimental results related to two types of data poisoning attacks against a logistic regressor. 

%Online Learning involves the fake news detection model actively collecting news articles at different time instants. This ongoing data intake allows the model to remain up-to-date, by adapting to the ever-changing landscape of online information. It not only encompasses the new data but also integrates the historical data it has encountered.
%Within this dynamic learning framework, the deliberate introduction of poisoned samples may come into play. Indeed, some of these collected news articles can be intentionally manipulated to carry misleading information, with the effect of challenging the model's resilience and its ability to distinguish between genuine and fake news.
%Figure~\ref{fig:process} describes the details of the process. More in detail, at each time $t$, previousely unseen news are combined with already existing ones; then, some poisoned data are generated and added to the collected data. Finally, the model is trained and updated using such set of data. 

%\begin{itemize}
 %   \item \textbf{A}: Over time, new news are produces and added to the existing news ($D^{t-1} \rightarrow D^t$).
 %   \item \textbf{B}: Poisoned data is generated and injected into the existing data at time $t$.    
 %   \item \textbf{C}: A subset of the data is collected at time $t$ and added to the previous time's data ($d^{t-1}$), forming $d^t$.   
%    \item \textbf{D}: The previous model $f^{t-1}$ is updated with the newly collected data, becoming $f^t$.
%\end{itemize}

%\subsection{Preliminary results on Logistic Regression}

\section{Data poisoning attacks}\label{sec::attacks}

\looseness -1 Within the online learning framework introduced in the previous section, we conduct preliminary experiments to analyze the effect of data poisoning attacks on Logistic Regression (LR) models, which are employed by many advanced techniques as the primary classifiers after feature extraction.% through representation learning. 
%By initially examining the LR model, we lay the groundwork to analyse the behaviour of more sophisticated models.
Two LR models are considered:
\begin{itemize}
    \item \textbf{Linear LR}: $\textit{logit}(\hat p) = \beta_0 + \beta_1x$
    \item \textbf{Quadratic LR}: $\textit{logit}(\hat p) = \beta_0 + \beta_1x + \beta_2x^2$,
\end{itemize}
where $\hat p$ represents the probability of classifying sample $x$ as fake news, $\beta_i \in \mathbb{R}\;\forall i \in \{0,1,2\}$ are the coefficients of the LR, and $\textit{logit}(\hat p) = ln\left(\frac{\hat p}{1-\hat p}\right)$.

\looseness -1 The examined attack strategy involves the following data poisoning methods:
\begin{itemize}
    \item \emph{Most Confidence Mislabeling}: a sample that is confidently classified by the model is added to the training data with the flipped label. 
    \item \emph{Target Label Flipping}: a sample identical to the target sample except for the label, which is flipped, is added to the training data.
\end{itemize}

\section{Results and discussion}\label{sec::results}

\begin{figure}[t]
\hfill
\subfigure[Linear LR]{\includegraphics[width=.48\textwidth]{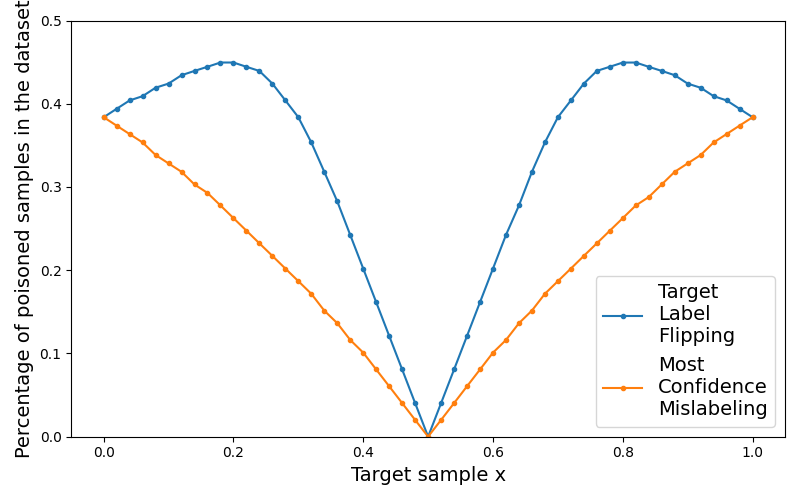}
\label{fig:poison1}
}
\hfill
\subfigure[Quadratic LR]{\includegraphics[width=.48\textwidth]{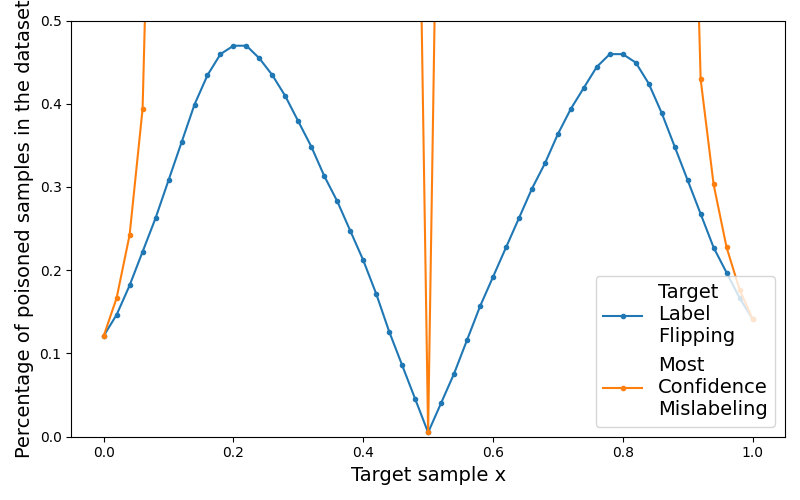}
\label{fig:poison2}
}
\hfill
\caption{Percentage of samples required to flip the target sample label, depending on its $x$ value and the type of poisoning used. The Most Confidence Mislabeling attack requires a lower number of samples to make the Linear LR model to misclassify the target news article. Conversely, the Quadratic LR model is resilient to it, but more significantly affected by the Target Label-Flipping attack.}
\label{fig:poison}
\end{figure}

To validate our theoretical framework, we propose experiments on synthetic data that allow us to show how these attacks can be performed on an online detector. In particular, these synthetic data will enable us to show how it is possible to manipulate the detector's behavior by appropriately modifying new training samples. The synthetic data used consists of $10000$ real values $x \in [0,1]$, with a binary class determined by the separation value $p=0.5$. % $p = (x \geq 0.5)$.} 
Figure~\ref{fig:poison} reports the comparison between the two poisoning attacks based on the number of samples required to misclassify the target sample.
%Our preliminary results demonstrate contrasting behaviors between the two LR models when subjected to different attack types.
Figure~\ref{fig:poison1} reveals that in the linear model, the Most Confidence Mislabeling attack requires a lower number of samples to misclassify the target news article. Conversely, Figure~\ref{fig:poison2} shows that the Quadratic LR model is resilient to Most Confidence Mislabeling, but more significantly affected by the Target Label-Flipping attack.

\begin{figure}[t]
\hfill
\subfigure[Most Confidence Mislabeling]{\includegraphics[width=.48\textwidth]{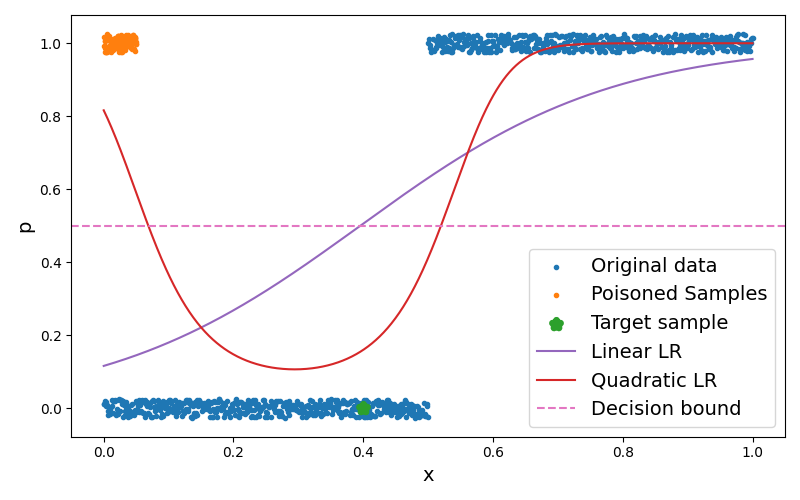}\label{fig:lr1}}
\hfill
\subfigure[Target Label Flipping]{\includegraphics[width=.48\textwidth]{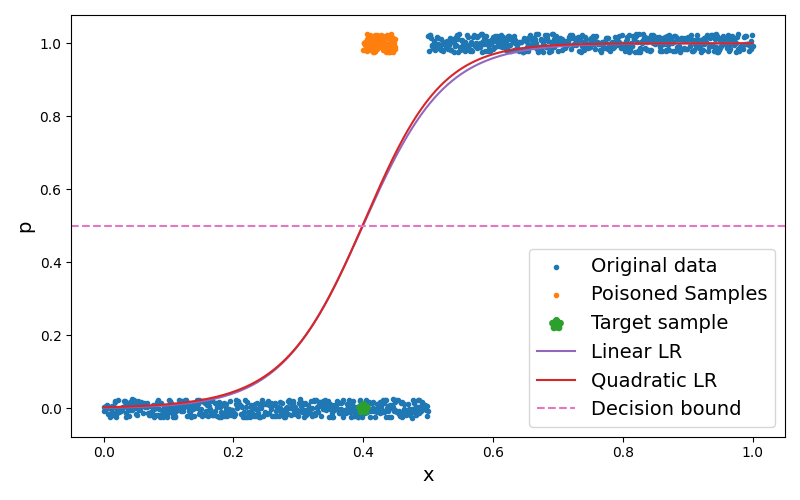}\label{fig:lr2}}
\hfill
\caption{$x$ value of the original data, the poisoned data, and the target sample. The two Logistic Regression models trained on the poisoned data are also displayed. When subjected to the Target Label Flipping poisoning, both models slightly alter their predictions to misclassify the target sample. In contrast, the Power Model adapts to the Most Confidence Sampling poisoning while maintaining the correct classification of the target sample.}
\label{fig:lr}
\end{figure}

\looseness-1 Figure~\ref{fig:lr} shows the $x$ value of the original data distribution, the chosen target sample, and $1000$ poisoned samples. The two LR models trained on the poisoned data are depicted as red and purple curves.
Figure~\ref{fig:lr1} shows that with the Most Confidence Mislabeling, the Quadratic LR model (the one with the most parameters) is able to follow the poisoned data and still correctly predict the target sample. This phenomenon occurs because the increased complexity of the Quadratic LR model allows it to adapt and capture patterns in the poisoned data, enabling accurate predictions of the target sample even in the presence of adversarial manipulation.
In contrast, in Figure~\ref{fig:lr2}, both models shift their decision bound and incorrectly classify the target sample when applying Target Label Flipping.

These initial findings highlight the importance of model architecture and complexity in determining their vulnerability to specific types of adversarial attacks. The Linear LR model exhibits greater susceptibility to Most Confidence Mislabeling attacks, while the Quadratic LR model demonstrates resistance to this attack type but remains vulnerable to Target Label Flipping. These results provide valuable insights into the behavior of LR models under adversarial data poisoning attacks, laying the foundation for further exploration of more sophisticated models and defense mechanisms.

\section{Conclusion and Future Work}\label{sec::concl}

In this paper, we delve into the uncharted territory of adversarial data poisoning attacks within the context of fake news detection. 
The proposed method formalizes an online learning framework where an online learner is pushed toward the misclassification of a true news article as false without any direct modification of the target article. 
Specifically, we introduced two types of data poisoning attacks, namely Most Confidence Mislabeling and Target Label Flipping, and we evaluated their impact on the performance of two logistic regression (LR). Results indicate that the susceptibility of the models varies on the basis of their complexity. It is important to remark that the effectiveness of data poisoning attacks may vary depending on the model architecture, the dataset used for training, and the robustness of the model.  

This work represents only preliminary efforts so far. In the future, we plan to train and evaluate several fake news detection models on real-world datasets which are commonly analysed in the field (e.g., PolitiFact\footnote{\url{https://www.politifact.com/}}, Gossipcop\footnote{\url{https://www.gossipcop.com/}}, FakeNewsNet~\cite{shu2018fakenewsnet}, Weibo21~\cite{nan2021mdfend}, FbMultiLingMisinfo~\cite{barnabo2022fbmultilingmisinfo}). Furthermore, the inclusion of various sources and types of misinformation will enable us to assess the robustness and generalizability of the proposed method across different contexts and sources of information. 
Another possible research direction is constituted by the accurate analysis of a model's performance if subject to other possible data poisoning attacks. For example, if the attacker has access to the information about the model gradient, a sample that maximizes the gradient on the target sample might be added to the training (e.g. Gradient Maximization).

Finally, a range of both traditional and deep learning models, which have shown promising performance in identifying fake news articles, can be considered. Possible examples of models include, among others, Support Vector Machines~\cite{liu2018early}, Convolutional Neural Networks~\cite{kim2014convolutional}, Transformer-based Models, Graph neural networks~\cite{nguyen2020fang}. A thorough examination of these models will be necessary to assess their susceptibility to data poisoning attacks. 

\printbibliography %Prints bibliography

\end{document}